\documentclass{article} % For LaTeX2e
\usepackage{baichuan}

\usepackage[utf8]{inputenc} % allow utf-8 input
\usepackage[T1]{fontenc}    % use 8-bit T1 fonts
\usepackage{hyperref}       % hyperlinks
\usepackage{url}            % simple URL typesetting
\usepackage{booktabs}       % professional-quality tables
\usepackage{amsfonts}       % blackboard math symbols
\usepackage{nicefrac}       % compact symbols for 1/2, etc.
\usepackage{microtype}      % microtypography
\usepackage{xcolor}         % colors
\usepackage{colortbl}
\usepackage{xcolor}
\definecolor{tOrange}{RGB}{210,105,30}
\definecolor{tBlue}{rgb}{0.39,0.58,0.93}
\definecolor{tPink}{RGB}{255,20,147}
\definecolor{tGreen}{RGB}{50,205,50}
\definecolor{tGold}{RGB}{255,239,210}

\definecolor{tPurple}{rgb}{0.95, 0.95, 1.0}
\definecolor{tGreen}{rgb}{0.95, 1.0, 0.95}
\definecolor{tRed}{rgb}{1.0, 0.95, 0.95}
\definecolor{colorOther}{rgb}{0.95, 0.95, 0.95}
\usepackage{multirow}
\usepackage{multicol}
\usepackage{amsmath}
\usepackage{graphicx}
\usepackage{amsmath}

\usepackage{subcaption}
\usepackage{utfsym}
\usepackage{svg}
\usepackage{array}
\usepackage{graphicx}
\usepackage{fancyhdr}
\usepackage{tablefootnote}
\usepackage{footnote}
\usepackage{CJKutf8}
\usepackage{graphicx}

\newcommand{\paratitle}[1]{\vspace{1.5ex}\noindent\textbf{#1}}
\newcommand{\ie}{\emph{i.e., }}

\newcommand{\ignore}[1]{}
\title{\centering \emph{Unveiling the Flaws:} Exploring Imperfections in Synthetic Data and Mitigation Strategies for Large Language Models}

\author{
	\begin{tabular}[t]{c}
		\quad\quad\quad Jie Chen\textsuperscript{1,2}$^{*}$, 
		Yupeng Zhang\textsuperscript{1}\thanks{Equal contribution}, 
		Bingning Wang\textsuperscript{1}$^{\dagger}$, 
		Wayne Xin Zhao\textsuperscript{2}\thanks{ Corresponding author, \texttt{\href{mailto:daniel@baichuan-inc.com}{daniel@baichuan-inc.com}},  
			\texttt{\href{mailto:batmanfly@gmail.com}{batmanfly@gmail.com}}},
		\\
		\quad\quad\quad Ji-Rong Wen\textsuperscript{2}, 
		and Weipeng Chen\textsuperscript{1}
		\\
		\\
		\quad\quad\quad\textnormal{\textsuperscript{1}Baichuan Inc.}
		\\
		\quad\quad\quad\textnormal{\textsuperscript{2}Gaoling School of Artificial Intelligence, Renmin University of China}
		\\
	\end{tabular}
}

\colmfinalcopy

\begin{document}

	\maketitle

	\begin{abstract}\label{abstract}
		
		Synthetic data has been proposed as a solution to address the issue of high-quality data scarcity in the training of large language models (LLMs). Studies have shown that synthetic data can effectively improve the performance of LLMs on downstream benchmarks. However, despite its potential benefits, our analysis suggests that there may be inherent flaws in synthetic data. The uniform format of synthetic data can lead to pattern overfitting and cause significant shifts in the output distribution, thereby reducing the model's instruction-following capabilities. Our work delves into these specific flaws associated with question-answer (Q-A) pairs, a prevalent type of synthetic data, and presents a method based on unlearning techniques to mitigate these flaws. The empirical results demonstrate the effectiveness of our approach, which can reverse the instruction-following issues caused by pattern overfitting without compromising performance on benchmarks at relatively low cost. Our work has yielded key insights into the effective use of synthetic data, aiming to promote more robust and efficient LLM training.
	\end{abstract}
	\section{Introduction}\label{introduction}
	The remarkable success of large language models (LLMs)~\citep{zhao2023survey} largely depends on the quality and diversity of the datasets used for training. However, acquiring large amounts of high-quality data can be challenging due to data scarcity, privacy concerns, and high costs~\citep{liu2024best}.
	% Some pessimists predict that it will run out of fresh text data in 2050~\citep{villalobos2022will}.
	Synthetic data has emerged as a promising solution to address these challenges~\citep{Sergey2019Synthetic}.
	
	% 什么是合成数据，合成数据为什么好，哪些方面取得了哪些效果？举几个例子
	Synthetic data, generated through algorithms or generative models rather than collected from real-world events, can be produced at scale and supplement areas where real-world data is scarce or difficult to obtain, such as in mathematical or reasoning tasks. Numerous studies have demonstrated the efficacy of synthetic data in improving model performance~\citep{microsoft2024phi,mukherjee2023orca}. Among the various methods of generating synthetic data, a common approach is the creation of synthetic question-answer (Q-A) pairs~\citep{Nemotron_4_340B_8T,maini2024rephrasing,wei2023magicoder}, as Q-A pairs exhibit diversity and richness, encompassing a range of question types from simple factual queries to complex reasoning problems. Another prevalent method is to generate data closely mimicking downstream tasks~\citep{luo2023wizardmath,yu2023metamath}. These methods have achieved excellent performance on both general-purpose and specialized benchmarks for LLMs.
	
	\begin{figure}[ht]
		\centering
		\includegraphics[width=1\textwidth]{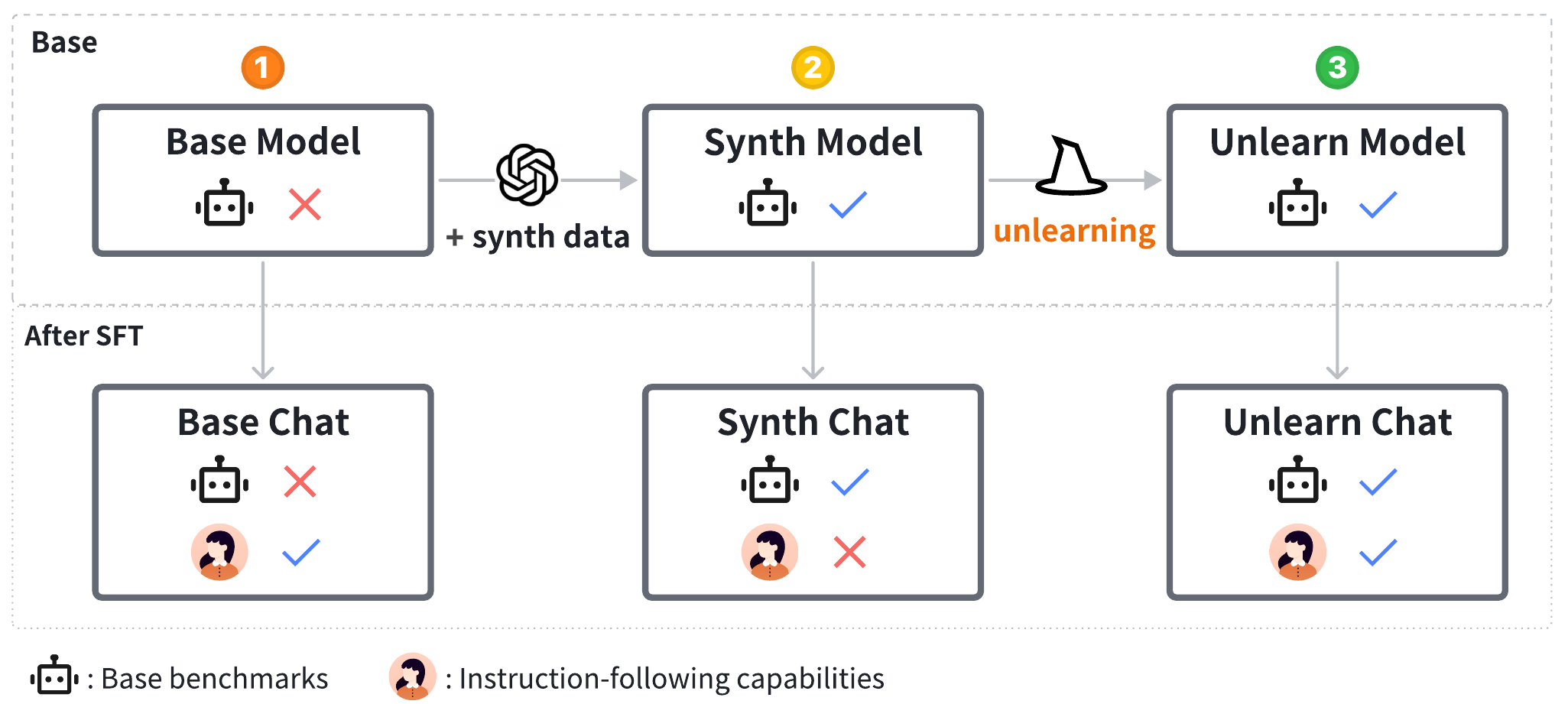}
		\caption{The overall pipeline of our study.}
		\label{pipeline}
	\end{figure}
	
	Despite numerous experiments demonstrating that synthetic data significantly enhances the capabilities of pre-trained models on downstream benchmarks, in this work, we observe a notable decline in the instruction-following capabilities of models after being pre-trained on synthetic data, specifically on synthetic Q-A pairs generated by GPT-4, and subsequent supervised fine-tuning (SFT). This observation prompts a deeper investigation into the underlying causes. While existing studies have extensively covered the applications of synthetic data, there is a notable lack of studies examining its impact on the instruction-following capabilities of LLMs. Furthermore, studies addressing the flaws in synthetic data have primarily focused on historical models or those with capabilities similar to currently trained models~\citep{shumailov2024curse,seddik2024bad,alemohammad2023selfconsuming}, leaving a gap in exploring the deficiencies of synthetic data generated by advanced models like GPT-4.
	
	Our work focuses on exploring the inherent flaws of synthetic data and its impact on LLMs. We find that the token distribution of synthetic data significantly differs from that of the real pre-training data, with synthetic data patterns being relatively uniform. Consequently, models trained on such synthetic data are likely to experience pattern overfitting, leading to substantial shifts in their output distributions and resulting in inferior performance.
	
	Based on these observations, we propose a novel strategy that leverages unlearning techniques to reduce the impact of misleading synthetic data patterns while preserving the LLM's foundational abilities on benchmarks and restoring its instruction-following capabilities. This strategy employs a lower-bounded forgetting loss, which is controllable and superior to traditional unlearning approaches. Our experimental results demonstrate that this strategy effectively mitigates the adverse impacts of synthetic data, balancing the LLM's performance on benchmarks with its ability to follow instructions at significantly low training costs. Our contributions are summarized as follows:
	
	$\bullet$ \textbf{Identification of Synthetic Data Limitations:} We provide a comprehensive analysis of the inherent limitations in synthetic data, specifically synthetic Q-A pairs, focusing on data distribution differences and pattern overfitting observed in models.
	
	$\bullet$ \textbf{Unlearn Method to Address Synthetic Data Issues:} We propose a novel unlearning strategy that effectively mitigates the adverse effects of synthetic data, thereby preserving the LLM's foundational abilities on benchmarks while reversing its instruction-following capabilities at significantly low training costs.
	
	\section{Related Work}
	
	\paratitle{Applications and Limitations of Synthetic Data.} Studies have shown that synthetic data has achieved remarkable results on downstream benchmarks~\citep{luo2023wizardmath,microsoft2024phi,mukherjee2023orca,wei2023magicoder}, addressing issues such as data scarcity and privacy~\citep{liu2024best,villalobos2022will,maini2024rephrasing}. For instance, Microsoft's Phi-3~\citep{microsoft2024phi} model, trained on heavily filtered publicly available web data and synthetic data, has outperformed much larger models on both academic benchmarks and internal testing. MagicoderS-CL-7B~\citep{wei2023magicoder}, a 7B parameter code model trained on synthetic code problems and answers generated by LLMs, even surpasses the prominent ChatGPT on many coding benchmarks. However, synthetic data is not without flaws. Several critical issues have been identified, particularly concerning model performance and data distribution integrity. One significant concern is the phenomenon of model collapse~\citep{shumailov2024curse,seddik2024bad}, 
	where training on model-generated data leads to the disappearance of the tails of the original content distribution. Furthermore, the recursive use of synthetic data in training generative models can amplify artifacts and biases, ultimately degrading model performance, as demonstrated by the concept of Model Autophagy Disorder (MAD)~\citep{alemohammad2023selfconsuming}. Task-specific synthetic data often lacks diversity and exhibits regional biases~\citep{yu2023large}, with effectiveness varying by task nature~\citep{li2023synthetic}.
	
	\paratitle{LLM Unlearning.} Unlearning in LLMs involves the elimination of specific undesired targets while preserving overall performance~\citep{liu2024rethinking}. Strategies vary from specific data points to higher-level concepts such as harmful language or specific knowledge domains~\citep{jang2022knowledge,lu2022quark,eldan2023whos}. Effective unlearning requires robustness and generalization~\citep{patil2023can,maini2024tofu,shi2023detecting} with efficient handling of computational costs~\citep{pawelczyk2023context}. Existing unlearning methods leverage various fine-tuning techniques, including gradient ascent, parameter-efficient fine-tuning, and KL-divergence-based methods, each with unique strengths and limitations regarding runtime and memory costs~\citep{yao2024large,jang2022knowledge,eldan2023whos}. While unlearning methods have been utilized to manage harmful data and reduce hallucinations in models, their application to synthetic data remains underexplored. Our research aims to fill this gap by applying unlearning strategies to mitigate the adverse effects of synthetic data on LLMs.
	
	\section{Experimental Setup}\label{exp_setup}
	
	In this section, we outline the experimental design, including dataset selection, model configurations, and evaluation benchmarks.
	
	\paratitle{Datasets.} We utilize five distinct datasets:
	
	$\bullet$~\textit{NonSynth data}: A comprehensive non-synthetic dataset collected from diverse sources~\citep{dolma,refinedweb,cerebras2023slimpajama}, including webpages, books, research papers, and codebases.
	
	$\bullet$~\textit{SynthQA data}: Synthetic Q-A pairs generated by GPT-4, based on a variety of sources including webpages, books, and other textual materials, covering topics such as mathematics, coding, and general knowledge.
	
	$\bullet$~\textit{MixedIns data}: Instructions consisting of general knowledge, mathematics, and coding, primarily generated by GPT-4 and human contributors.
	
	$\bullet$~\textit{U33B data}~\citep{yuan2023scaling}: Aggregated synthetic dataset of diverse reasoning paths generated from GSM8K dataset by multiple LLMs to enhance mathematical reasoning capabilities.
	
	$\bullet$~\textit{OpenHermes-2.5 data}~\citep{OpenHermes_2.5}: An extension of the OpenHermes-1 dataset, primarily consisting of synthetically generated instruction and chat samples.
	
	\paratitle{Models.} We use the following models in our experiments:
	
	\begin{table*}[t]
		\small
		\centering
		\begin{tabular}{lccccc}
			\toprule
			\textbf{Position Embedding} & \textbf{Hidden Size} & \textbf{FFN Size}& \textbf{Heads} & \textbf{Layers} & \textbf{Context Length}\\
			\midrule    
			RoPE~\citep{su2023roformer} & $2,048$  & $5,504$ & $32$ & $32$ & $4,096$  \\
			\bottomrule
		\end{tabular}
		\caption{The architecture details of BaseLM.}
		\label{tab:baselm}
		\vspace{-0.1cm}
	\end{table*}
	
	$\bullet$~\textit{BaseLM}: A Llama-like~\citep{touvron2023llama} 2B model trained from scratch. We set the learning rate to $1.0\times 10^{-4} $ and adopt a cosine learning rate schedule, training on a total of $1$ trillion tokens. The details of hyperparameters are listed in Table~\ref{tab:baselm}.
	
	$\bullet$~\textit{BaseLM-Chat (MixedIns/OpenHermes-2.5)}: Chat models obtained by performing SFT on BaseLM using MixedIns or OpenHermes-2.5 data. We set the learning rate to $2.0 \times 10^{-5}$, the number of epochs to $2$, the context length to $4,096$, and the batch size to $64$.
	
	\paratitle{Benchmarks.} We evaluate the capabilities of models using the following benchmarks:
	
	$\bullet$~\textit{Bilingual Capabilities}: Evaluated using the MMLU~\citep{hendrycks2021measuring}, CMMLU~\citep{li2024cmmlu} and C-Eval~\citep{huang2023ceval} benchmarks to assess the models' proficiency in handling both English and Chinese tasks.
	
	$\bullet$~\textit{Coding Proficiency}: Assessed with the HumanEval~\citep{chen2021evaluating} and MBPP~\citep{austin2021program} benchmarks, which measure the models' ability to generate correct and efficient code snippets based on given problems.
	
	$\bullet$~\textit{Mathematical Reasoning}: Measured using the GSM8K~\citep{cobbe2021training} benchmark, which tests the models' ability to solve complex mathematical problems.
	
	$\bullet$~\textit{Instruction-Following Capability}: Analyzed through FollowBench~\citep{jiang2024followbench} and MT-bench~\citep{zheng2023judging}, evaluating the models' ability to understand and follow complex instructions.
	
	\section{Defect Analysis of Synthetic Data}
	
	In this section, we systematically analyze the flaws of synthetic data, specifically synthetic Q-A pairs, by examining their data distribution differences and pattern overfitting observed in LLMs. This analysis is crucial to understand how synthetic data impacts the LLMs' foundational abilities on benchmarks and instruction-following capabilities.
	
	\subsection{Data Distribution Differences}\label{sec:data_dis}
	
	One of the primary concerns with synthetic data is the potential mismatch between its distribution and that of real-world data. This discrepancy can result in models that perform well on synthetic data but fail to generalize effectively to real-world scenarios.

	\begin{figure}[t]
		\centering
		\includegraphics[width=0.5\textwidth]{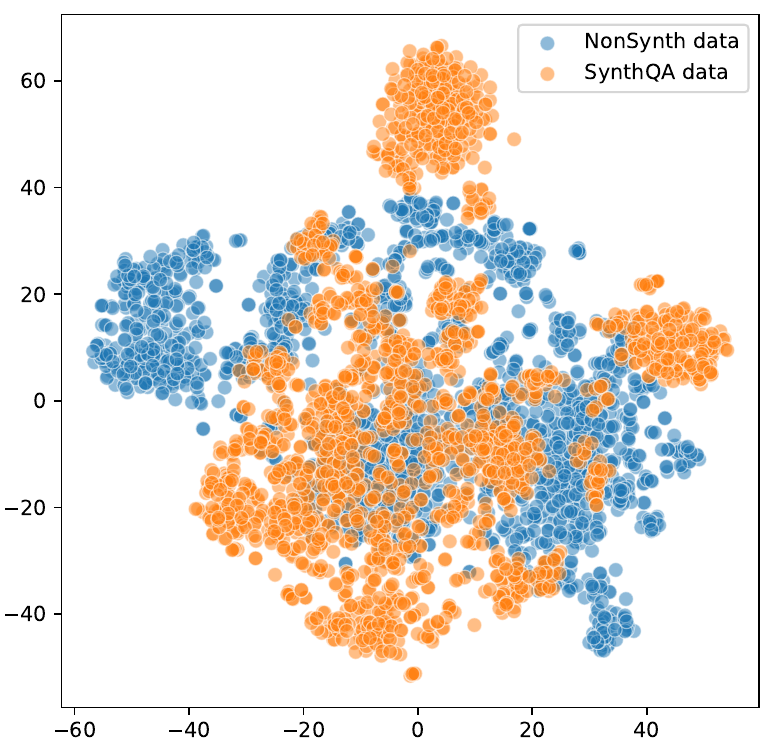}
		\caption{t-SNE visualization of data distributions. The clusters of NonSynth and SynthQA data show considerable non-overlap.}
		\vspace{-0.2cm}
		\label{fig:data_feature}
	\end{figure}
	
	\paratitle{Data Characteristic Differences.} Synthetic data generated by LLMs often exhibits distinct distributional characteristics compared to non-synthetic data. To illustrate these differences, we sample $2,000$ entries from both NonSynth and SynthQA data. Using the embeddings from the last hidden state of BaseLM, we apply t-SNE~\citep{van2008visualizing} for dimensionality reduction and visualize the data distributions in Figure~\ref{fig:data_feature}. The t-SNE visualization reveals that the clusters of NonSynth and SynthQA data have considerable areas of non-overlapping, which indicates that SynthQA data does not perfectly replicate the characteristics of NonSynth data. Such differences may lead to misinterpretations of real-world scenarios by LLMs trained on synthetic data.
	
	\begin{figure}[t]
		\centering
		\includegraphics[width=0.7\textwidth]{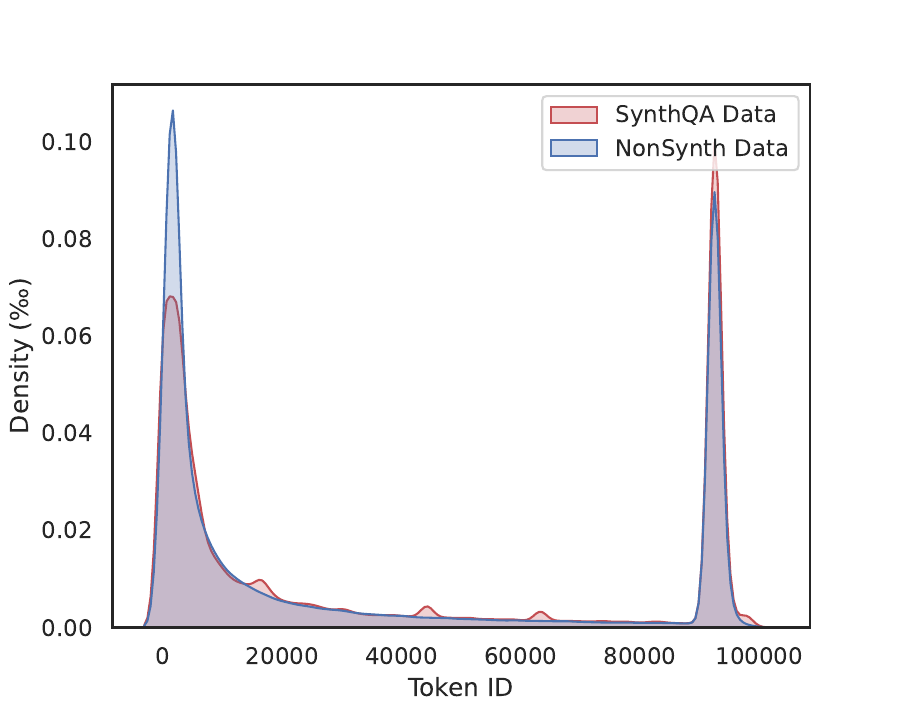}
		\caption{Kernel density estimation of token IDs for NonSynth and SynthQA data. The token frequency distribution for SynthQA data shows several small peaks, indicating high structural consistency for specific tokens compared to NonSynth data.}
		\label{fig:token_freq_density}
		\vspace{-0.2cm}
	\end{figure}
	
	\paratitle{Simplified Data Patterns.} Synthetic data often contains repetitive and structurally predictable elements, which simplify the complexity of real-world interactions and patterns. This simplification can result in data that fails to capture the intricacies of human language and interaction. To explore this, we again sample $2,000$ entries from both NonSynth and SynthQA data and calculate the token frequencies based on the tokenizer of BaseLM. Figure~\ref{fig:token_freq_density} presents the kernel density estimation (KDE)~\citep{parzen1962estimation} plot of token IDs. We observe that the distribution of token frequencies for SynthQA data exhibits several noticeable small peaks compared to NonSynth data. We find that these peaks correspond to tokens with a high degree of structural consistency within SynthQA data. Specifically, tokens like "question" (ID: 44246), "answer" (ID: 63264), and "summary" (ID: 16752) contribute to these observable peaks. The presence of these structural tokens indicates a repetitive pattern in SynthQA data, reflecting its inherent simplicity and lack of variability compared to NonSynth data. By over-representing certain tokens, synthetic datasets risk failing to encapsulate the full spectrum of linguistic diversity found in non-synthetic data, which may lead to models trained on such data being less robust and adaptable.
	
	\subsection{Pattern Overfitting}\label{sec:output-distribution}
	
	In this part, we investigate the detrimental effects of synthetic data on instruction-following capabilities and output distributions of LLMs. Our analysis highlights how synthetic data, specifically synthetic Q-A pairs, can cause overfitting to specific patterns observed in Section~\ref{sec:data_dis}, potentially affecting the performance of chat models.
	
	\begin{figure*}
		\begin{subfigure}[t]{0.47\textwidth}
			\centering
			\includegraphics[width=\textwidth]{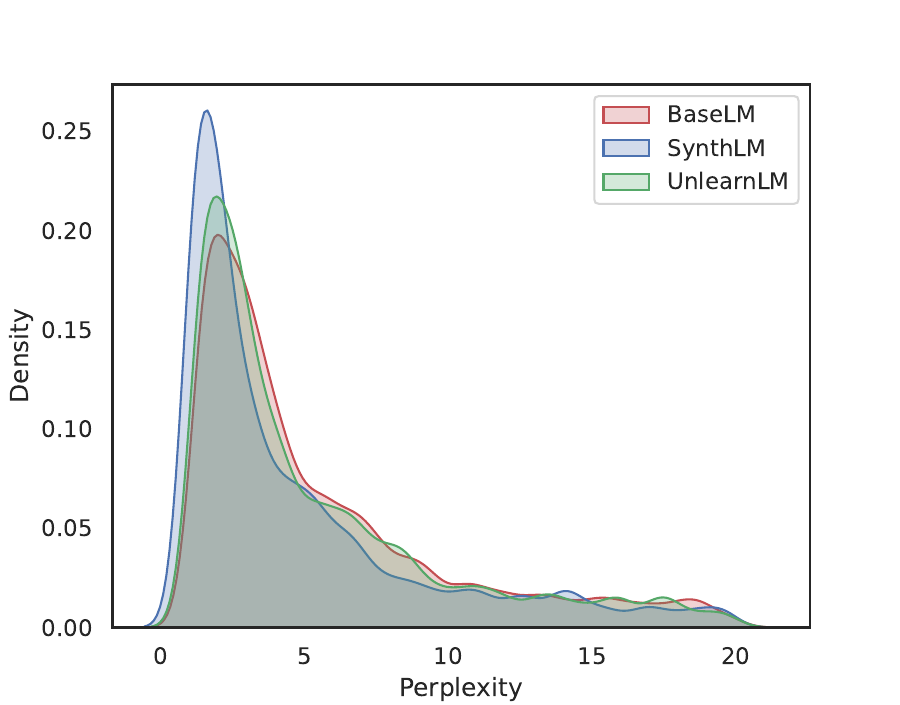}
			\caption{OpenHermes-2.5}
		\end{subfigure}	
		\begin{subfigure}[t]{0.47\textwidth}
			\centering
			\includegraphics[width=1\textwidth]{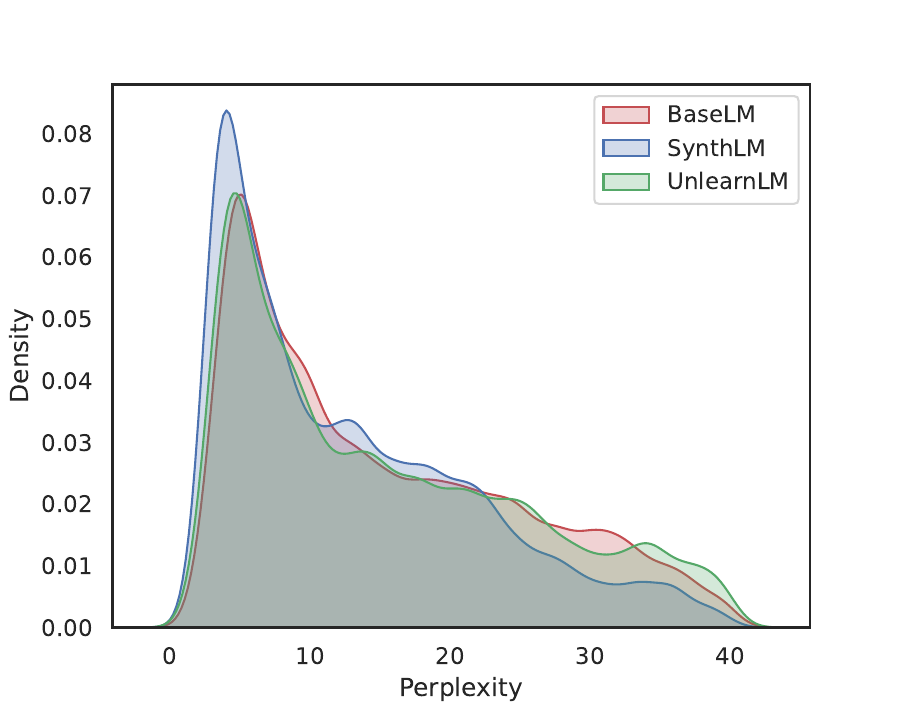}
			\caption{MixedIns}	
		\end{subfigure}	
		\caption{Kernel density estimation of perplexity values for OpenHermes-2.5 and MixedIns data using BaseLM, SynthLM and UnlearnLM. SynthLM shows a noticeable shift and reduced variance, while UnlearnLM corrects the distribution shift.}
		\label{fig:ppl_density}
		
	\end{figure*}
	
	\paratitle{Instruction-Following Capability Decline.} While synthetic data has shown considerable potential in enhancing the foundational abilities on benchmarks for LLMs in the pre-training stage, our work identifies significant challenges when these models undergo SFT. Specifically, we observe a notable decline in the instruction-following capabilities of chat models, underscoring critical limitations associated with the use of synthetic Q-A pairs. To investigate this issue, we design a series of experiments. We mix 2\% SynthQA data with NonSynth data to create a dataset containing $300$ billion tokens and perform continued pre-training on BaseLM with a fixed learning rate of $5.0 \times 10^{-5}$. The evaluation results, presented in Table~\ref{tab:base_bm} (\colorbox{tRed}{SynthLM} v.s. \colorbox{tPurple}{BaseLM}), show that the foundational abilities of BaseLM has significantly improved after training with synthetic Q-A pairs. We validate the role of synthetic data through ablation experiments in Section \ref{sec:ablation}. However, following SFT, we notice a severe decline in instruction-following capabilities in the resulting chat model, as shown in Table~\ref{tab:chat_bm} (\colorbox{tRed}{SynthModel-Chat} v.s. \colorbox{tPurple}{BaseLM-Chat}).
	
	\paratitle{Output Distribution Changes.} Due to simplified data patterns in synthetic data, a critical concern is its propensity to cause overfitting. To investigate this effect, we sample $2,000$ entries each from OpenHermes-2.5 and MixedIns data. We then calculate their perplexity using BaseLM and SynthLM. Figure~\ref{fig:ppl_density} shows the KDE plot of perplexity values for these two types of data. We can clearly observe that the perplexity distribution for SynthLM exhibits a noticeable shift and reduced variance compared to BaseLM, which is similar to the phenomenon of model collapse~\citep{shumailov2024curse}. This suggests a tendency for the model to overfit to the patterns present in the synthetic data, reducing its ability to deal with real-world variability.
	
	\section{Unlearning-Based Mitigation Strategy}
	
	In this section, we introduce our unlearning strategy and describe the experiments conducted to implement this approach.
	
	\subsection{Unlearning Strategy}\label{sec:unlearn_strategy}
	
	To address the identified flaws in synthetic data, we propose a mitigation strategy based on unlearning techniques. Typically, unlearning is applied to remove harmful data or reduce model hallucinations. In this context, we leverage unlearning to recalibrate the LLM's understanding, mitigating the adverse effects of synthetic data while preserving its beneficial attributes.
	
	\paratitle{Task Description.} In the task where the LLM predicts the next token \( y_i \) based on an existing token sequence \( y_{<i} = [y_1, y_2, \ldots, y_{i-1}] \), let \( p(y_{<i}; \theta) \) denote the predicted probability of \( y_i \). Formally, this can be expressed as:
	\[ p(y_{<i}; \theta) = P(y_i \mid y_{<i}; \theta), \]
	where \( \theta \) represents the parameters of the LLM. The prediction accuracy is evaluated using the cross-entropy loss function. Specifically, the loss for predicting \( y_i \) is given by $l(p(y_{<i}; \theta), y_i)$, where \( l(\text{input}, \text{target}) \) denotes the cross-entropy loss between the predicted probability distribution and the actual target token.
	
	\paratitle{Unlearning Loss.} Following previous work~\citep{yao2024large}, the unlearning loss function we designed consists of three parts:
	
	$\bullet$~\emph{Lower-Bounded Forgetting Loss}: This component focuses on forgetting the biased distribution of specific synthetic data. Unlike previous methods that apply gradient ascent~\citep{thudi2022unrolling} (i.e., adding a negative sign to the cross-entropy loss to introduce irrelevant elements into the predictions), we have observed that this method has uncontrolled loss due to the logarithm approaching zero without a lower bound. Therefore, we designed a simple yet effective lower-bounded forgetting loss by inverting the model prediction probabilities in the cross-entropy loss. This retains the original forgetting loss function's features while adding a lower bound (\ie 0). We validate the effectiveness of our forgetting loss approach through ablation experiments in Section~\ref{sec:ablation}. The designed lower-bounded forgetting loss \( L_{\text{fgt}} \) can be defined as:
	\[ L_{\text{fgt}} = \sum_{i=1}^{|y^{\text{syn}}|} l(1 - p(y^{\text{syn}}_{<i}; \theta), y^{\text{syn}}_i). \]
	
	% The differences between these two losses are shown in the model structure diagram. Additionally, we compare the performance differences between these two methods in detail in the ablation experiments.
	
	$\bullet$~\emph{Replay Loss}: We sample a portion of the data from the trained non-specific synthetic data for replay, using the cross-entropy loss to allow the model to retain memory of historical knowledge. The replay loss \( L_{\text{rpy}} \) can be defined as:
	\[ L_{\text{rpy}} = \sum_{i=1}^{|y^{\text{non-syn}}|} l(p(y^{\text{non-syn}}_{<i}; \theta), y^{\text{non-syn}}_i). \]
	
	$\bullet$~\emph{Bias Mitigation Loss}: After unlearning, we aim to ensure that the LLM's output distribution on the trained non-specific synthetic data does not change excessively. Therefore, we calculate the KL divergence between the current model and the original model on the data used for replay, as the bias mitigation loss \( L_{\text{mtn}} \) to preserve the original performance:
	\begin{equation}
		\begin{split}
			L_{mtn} = \sum_{i=1}^{|y^{\text{non-syn}}|} 
			\text{KL}(p(y^{\text{non-syn}}_{<i}; \theta_{\text{ori}}) \\
			\parallel p(y^{\text{non-syn}}_{<i}; \theta_i)),
		\end{split}
	\end{equation}
	where \( \theta_{\text{ori}} \) represents the parameters of the original model.
	
	\noindent Finally, we obtain the total unlearning loss function as follows:
	\[ L_{\text{unlearn}} = w_{\text{fgt}} \cdot L_{\text{fgt}} + w_{\text{rpy}} \cdot L_{\text{rpy}} + w_{\text{mtn}} \cdot L_{\text{mtn}}, \]
	where \( w_* \) denotes the weights corresponding to each part of the loss \( L_* \).
	
	% % draft
	% The figure illustrates the process of the model's unlearning strategy.

	\subsection{Unlearning Experiments}\label{sec:unlearn_exp}
	
	In this part, we detail the experimental process of applying unlearning techniques. Our objective is mitigate the adverse effects on models trained with synthetic data. Specifically, we aim to enhance the instruction-following capabilities of models while preserving their foundational abilities.
	
	\begin{table*}[t]
		
		\small
		\centering
		\begin{tabular}{l|cccccccc}
			\toprule
			\textbf{Models} & \textbf{C-Eval} & \textbf{CMMLU} & \textbf{MMLU} & \textbf{HumanEval} & \textbf{MBPP} & \textbf{GSM8K} & \textbf{Avg.} \\
			\midrule[0.5pt]
			\rowcolor{tPurple}
			BaseLM & 39.05 & 38.83 & 38.08 & 9.76 & 12.00 & 15.09 & 25.47 \\
			\rowcolor{tRed}
			SynthLM & 47.71 & 47.56 & 47.27 & 18.90 & 18.40 & 16.60 & 32.74 \\
			\midrule[0.5pt]
			\rowcolor{tGreen}
			RefineLM & 46.79 & 47.15 & 45.82 & 17.07 & 18.30 & 13.42 & 31.42 \\
			\rowcolor{tGold}
			UnlearnLM & 48.09 & 47.29 & 47.53 & 20.73 & 18.60 & 11.45 & 32.28 \\
			\bottomrule
		\end{tabular}
		\caption{Evaluation results of base models with continued pre-training and unlearning. \colorbox{tRed}{SynthLM} is obtained by training \colorbox{tPurple}{BaseLM} with a dataset containing $300$ billion tokens, of which 2\% are from the SynthQA data. \colorbox{tGreen}{RefineLM} is derived from \colorbox{tRed}{SynthLM} by further training with an additional $300$ billion tokens of NonSynth data. \colorbox{tGold}{UnlearnLM} is obtained by performing our unlearning strategy on \colorbox{tRed}{SynthLM} using $1$ billion tokens from the SynthQA data.}
		% \vspace{-0.1cm}
		\label{tab:base_bm}
	\end{table*}
	\begin{table*}[t]
		\small
		%\hspace*{-1.5cm}
		\centering
		\resizebox{\textwidth}{!}{
			\begin{tabular}{l|cc|c|ccccccc}
				\toprule
				& \multicolumn{2}{c|}{\textbf{FollowBench}} & \multirow{-2}{*}{\centering\textbf{}} \\
				\cmidrule(lr){2-3}
				\multirow{-2}{*}{\centering\textbf{Models}}
				& \textbf{SSR} & \textbf{HSR}
				& \multirow{-2}{*}{\centering\textbf{MT-Bench}}
				& \multirow{-2}{*}{\centering\textbf{C-Eval}} & \multirow{-2}{*}{\centering\textbf{CMMLU}} & \multirow{-2}{*}{\centering\textbf{MMLU}} & \multirow{-2}{*}{\centering\textbf{HumanEval}} & \multirow{-2}{*}{\centering\textbf{MBPP}} & \multirow{-2}{*}{\centering\textbf{GSM8K}} \\
				\midrule[0.5pt]
				\rowcolor{tPurple}
				BaseLM-Chat & 39.95 & 27.58 & 5.45 & 39.92 & 40.16 & 41.55 & 18.29 & 17.80 & 14.33 \\
				\rowcolor{tRed}
				SynthLM-Chat & 38.29 & 24.00 & 5.39 & 49.50 & 48.37 & 49.06 & 21.95 & 22.60 & 22.21 \\
				\midrule[0.5pt]
				\rowcolor{tGreen}
				RefineLM-Chat & 39.60 & 25.22 & 5.43 & 47.71 & 47.40 & 47.08 & 17.68 & 23.60 & 22.37 \\
				\rowcolor{tGold}
				UnlearnLM-Chat & 42.00 & 27.87 & 5.85 & 49.12 & 48.83 & 48.82 & 20.12 & 21.80 & 21.99 \\
				\bottomrule
			\end{tabular}
		}
		\caption{Evaluation results of chat models with continued pre-training and unlearning. Models with the suffix "-Chat" represent chat models derived from their corresponding base models in Table~\ref{tab:base_bm} through SFT on the MixedIns data.}
		\label{tab:chat_bm}
		\vspace{-0.2cm}
	\end{table*}
	
	\paratitle{Basic Implementation.} We utilize NonSynth data containing $300$ billion tokens to perform continued pre-training on SynthLM in Table~\ref{tab:base_bm}, with the aim of recovering the model's instruction-following capabilities. We utilize a fixed learning rate of $5.0 \times 10^{-5}$ during the training process. From the results in Table~\ref{tab:base_bm} and~\ref{tab:chat_bm}, we can clearly observe that extensive training with non-synthetic data leads to enhanced instruction-following capabilities (\colorbox{tGreen}{RefineLM-Chat} v.s. \colorbox{tRed}{SynthLM-Chat}) at the cost of a decline in overall base model performance (\colorbox{tGreen}{RefineLM} v.s. \colorbox{tRed}{SynthLM}). However, this approach does not completely eliminate the negative impact of the synthetic data on the model.
	
	\paratitle{Unlearning Strategy Implementation.} We propose employing the unlearning strategy on SynthLM. We apply lower-bounded forgetting loss on texts from the SynthQA data with $1$ billion tokens. Concurrently, we perform replay loss and bias mitigation loss on the trained NonSynth data alongside the unlearning process. We use a fixed learning rate of $5.0 \times 10^{-5}$ and set the weights \( w_{\text{fgt}} = 0.01 \), \( w_{\text{rpy}} = w_{\text{mtn}} = 1 \). As can be seen from Table~\ref{tab:base_bm} and~\ref{tab:chat_bm}, although unlearning leads to a slight decrease in foundational abilities of base (\colorbox{tGold}{UnlearnLM} v.s. \colorbox{tRed}{SynthLM}) and chat (\colorbox{tGold}{UnlearnLM-Chat} v.s. \colorbox{tRed}{SynthLM-Chat}) models, especially math abilities, there is a considerable improvement in instruction-following capabilities (\colorbox{tGold}{UnlearnLM-Chat} v.s. \colorbox{tPurple}{BaseLM-Chat}).
	
	\paratitle{Distribution Shift Correction.} The unlearning process partially corrects the output distribution shift of the LLM. Following the experiments in Section~\ref{sec:output-distribution}, we include the perplexity distribution of UnlearnLM on OpenHermes-2.5 and MixedIns data in Figure~\ref{fig:ppl_density}. It can be observed that the distribution shift has been effectively corrected after unlearning, indicating a significant reduction in pattern overfitting.
	
	\noindent It's worth noting that the instruction-following capabilities of UnlearnLM-Chat after unlearning with just $\mathbf{1}$ \textbf{billion tokens} surpass the performance of both RefineLM-Chat trained on $\mathbf{300}$ \textbf{billion tokens} and BaseLM-Chat. Additionally, the foundational abilities of UnlearnLM are comparable to those of RefineLM, suggesting that the beneficial effects of synthetic data on model performance have been preserved.
	This underscores the efficacy of our method in achieving \textbf{more robust and efficient LLM training at significantly lower training costs}.
	
	\section{Ablation Study}
	\label{sec:ablation}
	
	\begin{table*}[!h]
		\small
		\centering
		\begin{tabular}{l|cc|c|c}
			\toprule
			& \multicolumn{2}{c|}{\textbf{FollowBench}} & \multirow{-2}{*}{\centering\textbf{}} \\
			\cmidrule(lr){2-3}
			\multirow{-2}{*}{\centering\textbf{Models}}
			& \textbf{SSR} & \textbf{HSR}
			& \multirow{-2}{*}{\centering\textbf{MT-Bench}}
			& \multirow{-2}{*}{\centering\textbf{GSM8K}} \\
			\midrule[0.5pt]
			BaseLM-Chat (O.H.) & 40.25 & 27.27 & 5.76 & 34.27 \\
			SynthLM* (U33B)-Chat (O.H.) & 39.95 & 25.13 & 5.61 & 43.06 \\
			UnlearnLM* (U33B)-Chat (O.H.) & 40.21 & 27.26 & 5.87 & 42.00 \\
			\bottomrule
		\end{tabular}
		\caption{Evaluation results of chat models with continued pre-training on U33B data and subsequent unlearning. SynthLM*(U33B) is the base model trained with $40$ billion tokens including 2\% U33B data. UnlearnLM*(U33B) is derived from SynthLM*(U33B) by applying our unlearning strategy. Models with the suffix "-Chat(O.H.)" represent chat models derived from their corresponding base model through SFT on the OpenHermes-2.5 data.}
		\label{tab:unlearn_u33b}
		\vspace{-0.2cm}
	\end{table*}
	
	\begin{table*}[t]
		\small
		\centering
		\begin{tabular}{l|ccccccc}
			\toprule
			\textbf{Models} & \textbf{C-Eval} & \textbf{CMMLU} & \textbf{MMLU} & \textbf{HumanEval} & \textbf{MBPP} & \textbf{GSM8K} & \textbf{Avg.} \\
			\midrule[0.5pt]
			BaseLM & 39.05 & 38.83 & 38.08 & 9.76 & 12.00 & 15.09 & 25.47 \\
			MixSynthLM & 44.63 & 44.12 & 45.00 & 18.29 & 19.40 & 14.95 & 31.07 \\
			NonSynthLM & 42.33 & 40.46 & 40.88 & 18.29 & 17.80 & 12.21 & 28.66 \\
			\bottomrule
		\end{tabular}
		\caption{Evaluation results of BaseLM with continued pre-training on synthetic and non-synthetic data. MixSynthLM is BaseLM trained with $40$ billion tokens including 2\% SynthQA data. NonSynthLM is BaseLM trained with $40$ billion tokens of NonSynth data.}
		% \vspace{-0.1cm}
		\label{tab:syn_or_nonsyn}
	\end{table*}
	
	\begin{table*}[t]
		\small
		\centering
		\begin{tabular}{l|ccccccc}
			\toprule
			\textbf{Models} & \textbf{C-Eval} & \textbf{CMMLU} & \textbf{MMLU} & \textbf{HumanEval} & \textbf{MBPP} & \textbf{GSM8K} & \textbf{Avg.} \\
			\midrule[0.5pt]
			SynthLM & 47.71 & 47.56 & 47.27 & 18.90 & 18.40 & 16.60 & 32.74 \\
			UnlearnLM (GA) & 26.58 & 25.08 & 39.28 & 11.59 & 9.60 & 6.82 & 19.82 \\
			UnlearnLM (\textbf{Ours}) & 48.09 & 47.29 & 47.53 & 20.73 & 18.60 & 11.45 & 32.28 \\
			\bottomrule
		\end{tabular}
		\caption{Evaluation results of SynthLM with different unlearning strategies applied. UnlearnLM (GA) is derived from SynthLM by applying traditional gradient ascent loss. UnlearnLM (Ours) is derived by applying our lower-bounded forgetting loss.}
		\vspace{-0.2cm}
		\label{tab:log_p_or_log1_p}
	\end{table*}
	
	\subsection{Effectiveness of Unlearning Strategy}
	
	To explore the effectiveness of our unlearning strategy across different types of synthetic data, we conduct experiments using the U33B data. We first perform continued pre-training on the BaseLM with $40$ billion tokens of data, including 2\% U33B data, resulting in SynthLM*(U33B). We utilize a fixed learning rate of $5.0 \times 10^{-5}$ during the training process. Following this, we apply our unlearning strategy to mitigate the adverse effects of U33B data on instruction-following capabilities while preserving its positive impact on foundational abilities, particularly in mathematics. Specifically, we employ the same unlearning parameters as described in Section~\ref{sec:unlearn_exp}, resulting in UnlearnLM*(U33B). We conduct SFT on the resulting models using OpenHermes-2.5 data. The evaluation results are presented in Table~\ref{tab:unlearn_u33b}. The results indicate that while the model trained with U33B data improves its mathematical abilities, it exhibits a decline in instruction-following capabilities. However, after applying our unlearning strategy, the instruction-following capabilities are restored, while retaining the enhancements in mathematical abilities provided by the U33B data. These findings suggest that our unlearning strategy could be extended to other types of open-source synthetic data.
	
	\subsection{Impact of Synthetic Data on Model Performance}
	
	To verify that SynthQA data, rather than NonSynth data, contributes to the significant performance improvements in BaseLM, we conduct a controlled ablation experiment. We evaluate two models: NonSynthLM, which is the BaseLM trained with $40$ billion tokens of NonSynth data, and MixSynthLM, which is the BaseLM trained with $40$ billion tokens of data including 2\% SynthQA data. To ensure a fair comparison and better verify the impact of synthetic data, the NonSynth data used to train both NonSynthLM and MixSynthLM is the same high-quality data corpus used to generate the SynthQA data. The evaluation result is shown in Table~\ref{tab:syn_or_nonsyn}. We can see that MixSynthLM exhibits markedly superior performance enhancements. This confirms that synthetic data plays a critical role in boosting base model performance.
	
	\subsection{Efficacy of Bounded Forgetting Loss}
	
	When introducing our unlearning strategy in Section~\ref{sec:unlearn_strategy}, we use the lower-bounded forgetting loss to forget the biased distribution of specific synthetic data. To evaluate the effectiveness of this approach compared to the traditional gradient ascent loss, we conduct a comparative experiment where the SynthLM in Table~\ref{tab:base_bm} undergo unlearning using both the lower-bounded forgetting loss and the traditional gradient ascent loss. As shown in Table~\ref{tab:log_p_or_log1_p}, we can clearly observe that the model subjected to traditional gradient ascent loss exhibits severe performance degradation. This may be due to the uncontrolled magnitude of negative loss during training. Conversely, the lower-bounded forgetting loss results only in a partial decline in mathematical abilities.
	
	\section{Conclusion}
	
	In this work, we have systematically explored the potential issues associated with synthetic data, particularly focusing on synthetic Q-A pairs, and their impact on the performance of LLMs. Our analysis has identified inherent flaws in synthetic data, such as pattern overfitting and significant shifts in output distribution, which can degrade the instruction-following capabilities of LLMs. To mitigate these adverse effects, we have proposed an innovative unlearning-based strategy. This strategy employs a lower-bounded forgetting loss, which is controllable and superior to traditional unlearning approaches at significantly lower training costs. The empirical results demonstrate that our strategy effectively addresses the limitations of synthetic data and corrects the output distribution shift, thereby enhancing the instruction-following capabilities while preserving foundational capabilities of LLMs on benchmarks. Our work has demonstrated a viable path to leverage the advantages of synthetic data without being adversely affected by its limitations, enhancing the robustness and efficiency of LLM training.
	
	\section{Limitations}
	
	Despite our substantial efforts, several limitations warrant further consideration. Firstly, while our unlearning-based strategy has shown promise in mitigating the negative effects of synthetic data, it may still cause degradation in specific model capabilities, such as mathematical reasoning. Moreover, its scalability to much larger models remains untested. As LLMs continue to grow in size and complexity, the computational efficiency and practical applicability of this strategy require further validation. Additionally, this study primarily focuses on the flaws and mitigation strategies related to Q-A pair synthetic data. Although we have demonstrated the effectiveness of our unlearning strategy on the open-source synthetic dataset U33B, many other forms of synthetic data remain unexplored. Furthermore, the quality of synthetic data generated by GPT-4 used in this study may not fully represent the entire spectrum of synthetic data quality. Different synthetic data generation techniques and tools can produce data with varying degrees of imperfections, potentially impacting the effectiveness of our mitigation strategy. Further investigation into more advanced unlearning techniques is necessary to minimize these side effects. We will continue to refine and enhance our method in future work.
	
	%%%%%%%%%%%%%%%%%%%%%%%%%%%%%%%%%%%%%%%%%%%%%%%%%%%%%%%%%%%%
	%		\clearpage
	%		\newpage
	\bibliography{baichuan}
	\bibliographystyle{baichuan}
	
	\newpage
	\appendix
	\section{Examples of Data Utilized in This Work}
	
	In Section~\ref{exp_setup}, we introduce the various datasets employed in our research. To provide a clear understanding of the data characteristics and content diversity, we present examples for each dataset type in Table~\ref{tab:data_example}.
	
	\begin{table*}[t]
		\small
		\centering
		\vspace{-20pt}
		\hspace*{-1.2cm}
		\begin{tabular}{lp{2.0cm}p{1.0cm}p{8.5cm}}
			\toprule
			\rowcolor[gray]{.92} \textbf{Dataset} & \textbf{Type} & \multicolumn{1}{l}{\textbf{Source}} & \multicolumn{1}{c}{\textbf{Sample}} \\
			\midrule
			\multirow{18}{*}{\textbf{NonSynth}} & \multirow{18}{*}{\begin{tabular}[x]{@{}l@{}}Non-synthetic\\ Data\end{tabular}} & \multirow{3}{*}{Webpage}                                                     & As an independent Nissan service repair centre and our aim is to provide our customers with an alternative to the high servicing and repair costs associated with large Nissan dealerships. \\
			\cmidrule{3-4}
			& & \multirow{4}{*}{Book} & The sovereign or heir of Moscow was to succeed Yan Kazimir, details of boundaries and succession to be settled by the Diet, both sides to refrain from hostilities till the Swedes were expelled, and neither to make peace with Sweden separately. \\
			\cmidrule{3-4}
			& & \multirow{5}{*}{\begin{tabular}[x]{@{}l@{}}Research \\ Paper\end{tabular}} & $\backslash$section*{Introduction}$\backslash$n$\backslash$nClubbing is a central part of many young adults' lives. In Norway, the club culture is alcohol driven and drinking to intoxication is a common phenomenon in the Nordic countries (Mäkelä et al., 2001), as in other parts of the Western world (Measham $\backslash\backslash$\& Brain, 2005; Moore, 2010). \\
			\cmidrule{3-4}
			& & \multirow{6}{*}{Codebase} & from openpyxl import load\_workbook$\backslash$nimport numpy as np $\backslash$n$\backslash$ndef read():$\backslash$n    \#Load data from workbook$\backslash$n    wb = load\_workbook('DL03\_Teste01\_Dados.xlsx')$\backslash$n    sheet = wb['Planilha1']$\backslash$n$\backslash$n    \#Recover data$\backslash$n    datasheet = []$\backslash$n    for row in sheet.iter\_rows():  $\backslash$n        newRow = list()$\backslash$n        for cell in row:$\backslash$n            newRow.append(cell.value) \\
			\midrule
			\multirow{14}{*}{\textbf{SynthQA}}        & \multirow{14}{*}{\begin{tabular}[x]{@{}l@{}}Synthetic\\ Q-A Pair\end{tabular}}  & \multirow{6}{*}{Mathematic}                                              & Question:$\backslash$n$\backslash$nWhich of the following represents the derivative of the function $f(x) = x^2$?$\backslash$n$\backslash$nA) 2$\backslash$n$\backslash$nB) 2x$\backslash$n$\backslash$nC) $x^2$$\backslash$n$\backslash$nD) $x^3$$\backslash$n$\backslash$n**Answer:**$\backslash$n$\backslash$nTo find the derivative of $f(x) = x^2$, we apply the power rule of differentiation. If $f(x) = x^n$, then $f'(x) = nx^{n-1}$.$\backslash$n$\backslash$nApplying the rule to $f(x) = x^2$, we have:$\backslash$n$\backslash$n $f'(x) = 2x^{2-1} = 2x\backslash$n$\backslash$nSo, the correct answer is **B) 2x**. \\
			\cmidrule{3-4}
			& & \multirow{4}{*}{Coding}                                                      & Create a test suite of quantum physics questions categorized by difficulty.$\backslash\backslash$n$\backslash$nBest Answer:$\backslash$nGiven the complexity, this task would typically be broken into multiple stages and require a team of experts ... if \_\_name\_\_ == "\_\_main\_\_":$\backslash$n    app.run() \\
			\cmidrule{3-4}
			& & \multirow{3}{*}{\begin{tabular}[x]{@{}l@{}}General\\ Knowledge\end{tabular}} & \begin{CJK}{UTF8}{gbsn}探讨迷茫心理、求助的重要性与技巧、心理学视角下的人际交流和资源交换原则。$\backslash$n从心理学来讲,人生之所以迷茫,是对未知的无能为力。 ... 明确对方需求,才能获得你想要的帮助。\end{CJK} \\
			\midrule
			\multirow{16}{*}{\textbf{MixedIns}} & \multirow{16}{*}{Instruction} & \multirow{4}{*}{Mathematic} & \begin{CJK}{UTF8}{gbsn}\{"role": "user", "content": "帮助用户解决以下数学问题。$\backslash$n问题: 计算 45 ÷ 5 + 6 × 2"\}, \{"role": "assistant", "content": "根据算术运算法则, 先计算乘除再计算加减：$\backslash$n45 ÷ 5 + 6 × 2 = 9 + 12 = 21"\}\end{CJK} \\
			\cmidrule{3-4}
			& & \multirow{6}{*}{Coding} & \begin{CJK}{UTF8}{gbsn}\{"role": "user", "content": "创建一个包含10个随机数字的列表，并按升序排列。$\backslash$n"\}, \{"role": "assistant", "content": "以下是Python语言的代码实现：$\backslash$n```python$\backslash$nimport random$\backslash$n\# 生成一个包含10个随机数字的列表$\backslash$nmy\_list = [random.randint(1, 100) for i in range(10)]$\backslash$n\# 使用sort()方法按升序排列$\backslash$nmy\_list.sort()$\backslash$n\# 输出$\backslash$nprint(my\_list)"\}\end{CJK} \\
			\cmidrule{3-4}
			& & \multirow{4}{*}{\begin{tabular}[x]{@{}l@{}} General\\ Knowledge\end{tabular}} & \begin{CJK}{UTF8}{gbsn}\{"role": "user", "content": "给定以下两种动物的名字，请确定它们属于哪个动物大类；鲨鱼和海马。$\backslash$n鲨鱼和海马$\backslash$n"\}, \{"role": "assistant", "content": "鲨鱼属于鱼类，而海马属于海龙目，也就是海马目动物。"\}\end{CJK} \\
			\midrule
			\multirow{5}{*}{\textbf{U33B}} & \multirow{5}{*}{\begin{tabular}[x]{@{}l@{}}Synthetic\\ Q-A Pair\end{tabular}}  & \multirow{5}{*}{Mathematic} & \{"query": "Natalia sold clips to 48 of her friends in April, and then she sold half as many clips in May. How many clips did Natalia sell altogether in April and May?", "response": "Natalia sold 48/2 = <<48/2=24>>24 clips in May.$\backslash$nNatalia sold 48+24 = <<48+24=72>>72 clips altogether in April and May.$\backslash$n\#\#\#\# 72"\} \\
			\midrule
			\multirow{3}{*}{\textbf{OpenHermes-2.5}} & \multicolumn{2}{c}{\multirow{3}{*}{Instruction and Chat Example}} & \{"from": "human", "value": "Do you know any jokes about animals with ailments?"\}, \{"from": "gpt", "value": "What's worse than a centipede with sore feet? A giraffe with a sore throat"\} \\
			\bottomrule
		\end{tabular}
		\caption{Illustrative examples for six datasets used in this work.}
		\label{tab:data_example}
	\end{table*}	
	
\end{document}